\documentclass[english,12pt]{article}
\usepackage[utf8]{inputenc}
\usepackage[top=3cm,bottom=3cm,left=3cm,right=3cm,marginparwidth=1.75cm]{geometry}

\usepackage{hyperref}
\hypersetup{
    colorlinks=true,
    linkcolor=blue,
    citecolor=brown,
}

\usepackage{amsmath,amsfonts,bm}

\def\eqref#1{equation~\ref{#1}}

\def\1{\bm{1}}

\def\ry{{\textnormal{y}}}

\DeclareMathAlphabet{\mathsfit}{\encodingdefault}{\sfdefault}{m}{sl}
\SetMathAlphabet{\mathsfit}{bold}{\encodingdefault}{\sfdefault}{bx}{n}

\newcommand{\E}{\mathbb{E}}

\newcommand{\R}{\mathbb{R}}

\newcommand{\Var}{\mathrm{Var}}

\usepackage{url}            
\usepackage{booktabs}       
\usepackage{amsfonts}       
\usepackage{nicefrac}       
\usepackage{microtype}      
\usepackage{xcolor}         
\usepackage{paralist}
\usepackage{wrapfig}
\usepackage{graphicx}
\usepackage{caption}
\usepackage{subcaption}
\newcommand{\norm}[1]{\left\lVert#1\right\rVert}
\usepackage{multirow}
\newcommand{\B}{\boldsymbol}

\newcommand{\C}{\mathcal}
\usepackage{bm}
\usepackage{pifont}

\usepackage{natbib}

\makeatletter  
\usepackage{stmaryrd}  
\newcommand{\fixed@sra}{$\vrule height 2\fontdimen22\textfont2 width 0pt\shortrightarrow$}
\newcommand{\shortarrow}[1]{%
  \mathrel{\text{\rotatebox[origin=c]{\numexpr#1*45}{\fixed@sra}}}
}
\makeatother

\begin{document}



\title{Flexible Modeling and Multitask Learning using Differentiable Tree Ensembles}

\author{
Shibal Ibrahim\thanks{MIT Department of Electrical Engineering and Computer Science. email: {\texttt{shibal@mit.edu}}}\and
Hussein Hazimeh\thanks{MIT Operations Research Center. email: {\texttt{hazimeh@mit.edu}}}\and  
Rahul Mazumder\thanks{MIT Sloan School of Management, Operations Research Center and MIT Center for Statistics. email: {\texttt{rahulmaz@mit.edu}}}
}

\date{August, 2021}

\maketitle

\begin{abstract}
Decision tree ensembles are widely used and competitive learning models. Despite their success, popular toolkits for learning tree ensembles have limited modeling capabilities. For instance, these toolkits support a limited number of loss functions and are restricted to single task learning. We propose a flexible framework for learning tree ensembles, which goes beyond existing toolkits to support arbitrary loss functions, missing responses, and multi-task learning.
Our framework builds on differentiable (a.k.a. soft) tree ensembles, which can be trained using first-order methods. However, unlike classical trees, differentiable trees are difficult to scale. We therefore propose a novel tensor-based formulation of differentiable trees that allows for efficient vectorization on GPUs. 
We perform experiments on a collection of 28 real open-source and proprietary datasets, which demonstrate that our framework
can lead to 100x more compact and 23\% more expressive tree ensembles than those by popular toolkits.
\end{abstract}

\section{Introduction}
Decision tree ensembles are popular models that have proven successful in various machine learning applications and competitions  \citep{Erdman2016, xgboost2016}. 
Besides their competitive performance, decision trees are appealing in practice because of their interpretability, robustness to outliers, and ease of tuning \citep{Hastie2009}. 
Training a decision tree naturally  requires solving a combinatorial optimization problem, which is difficult to scale. In practice, greedy heuristics are commonly used to get feasible solutions to the combinatorial problem; for example CART \citep{Breiman1984}, C5.0 \citep{Quinlan1993}, and OC1 \citep{Murthy1994}. By building on these heuristics, highly scalable toolkits for learning tree ensembles have been developed, e.g., XGBoost  \citep{xgboost2016} and LightGBM \citep{lightgbm2017}. These toolkits are considered a defacto standard for training tree ensembles and have demonstrated success in various domains. 

Despite their success, popular toolkits for learning tree ensembles lack modeling flexibility. For example, these toolkits support a limited set of loss functions, which may not be suitable for the application at hand. Moreover, these toolkits are limited to single task learning. In many modern applications, it is desired to solve multiple, related machine learning tasks. In such applications, multi-task learning, i.e., learning tasks simultaneously, may be a more appropriate choice than single task learning \citep{Chapelle2010, Kumar2012, Lei2015, Crawshaw2020}. If the tasks are sufficiently related, multi-task learning can boost predictive performance by leveraging task relationships during training.

In this paper, we propose a flexible modeling framework for training tree ensembles that addresses the aforementioned limitations. Specifically, our framework allows for training tree ensembles with any differentiable loss function, enabling  the user to seamlessly experiment with different loss functions and select what is suitable for the application. Moreover, our framework equips tree ensembles with the ability to perform multi-task learning. To achieve this flexibility, we build up on soft trees  \citep{Kontschieder2015, Hazimeh2020}, which are differentiable trees that can be trained with first-order (stochastic) gradient methods. Previously, soft tree ensembles have been predominantly explored for classification tasks with cross-entropy loss. In such cases, they were found to lead to more expressive and compact tree ensembles \citep{Hazimeh2020}. However, the state-of-the-art toolkits e.g., TEL \citep{Hazimeh2020}, are slow as they only support CPU training and are difficult to customize. Our proposed tree ensemble learning framework supports a collection of loss functions such as classification, regression, Poisson regression, zero-inflation models, overdispersed distributions, multi-task learning and has seamless support for others --- the user can modify the loss function with a single line.  
We show loss customization can lead to a significant reduction in ensemble sizes (up to 20x). We also propose a careful tensor-based formulation of differentiable tree ensembles, which leads to more efficient training (10x) on CPUs as well as support GPU-training (20x).

We propose a novel extension for multi-task learning with tree ensembles --- this may not be readily accommodated within popular gradient boosting toolkits. Our model can impose both 'soft' information sharing across tasks---soft information sharing is not possible with  popular multi-task tree ensemble toolkits e.g., RF \cite{Breiman2001}, GRF \cite{Athey2019}. Our proposed framework leads to $23\%$ performance gain (on average) and up to a 100x reduction in tree ensemble sizes on real-world datasets, offering a promising alternative approach for learning tree ensembles.

\paragraph{Contributions} Our contributions can be summarized as follows. 
\textbf{(i)} We propose a flexible framework for training differentiable tree ensembles with seamless support for new loss functions. 
\textbf{(ii)} We introduce a novel, tensor-based formulation for differentiable tree ensembles that allows for efficient training on GPUs. Existing toolkits e.g., TEL \cite{Hazimeh2020}, only support CPU training. 
\textbf{(iii)} We extend differentiable tree ensembles to multi-task learning settings by introducing a new regularizer that allows for soft parameter sharing across tasks.
\textbf{(iv)} We introduce a new toolkit (based on Tensorflow 2.0) and perform experiments on a collection of 28 open-source and real-world datasets, 
demonstrating that our framework
can lead to 100x more compact ensembles and up to $23\%$ improvement in out-of-sample performance, compared to tree  ensembles learnt by popular toolkits such as XGBoost  \cite{xgboost2016}.


\section{Related Work}
\label{sec:related-work}
Learning binary trees has been traditionally done in three ways. The first approach relies on greedy construction and/or optimization via methods such as CART \citep{Breiman1984}, C5.0 \citep{Quinlan1993}, OC1 \citep{Murthy1994}, TAO \citep{Carreira-Perpinan2018}. These methods optimize a criterion at the split nodes based on the samples routed to each of the nodes. The second approach considers probabilistic relaxations/decisions at the split nodes and performs end-to-end learning with first order methods \citep{Irsoy2012, Frosst2017, Lay2018}. The third approach considers optimal trees with mixed integer formulations and jointly optimize over all discrete/continuous parameters with MIP solvers \citep{Bennett1992, Bennett1996, Bertsimas2017, Zhu2020}.  
Each of the three approaches have their pros and cons. 
The first approach is highly scalable because of greedy heuristics. In many cases, the tree construction uses a splitting criterion different from the optimization objective \citep{Breiman1984} (e.g., gini criterion when performing classification) possibly resulting in sub-optimal performance. The second approach is also scalable but principled pruning in probabilistic trees remains an open research problem. The third approach scales to small datasets with samples $N \sim 10^4$, features $p \sim 10$ and tree depths $d \sim 4$.    

Jointly optimizing over an ensemble of classical decision trees is a hard combinatorial optimization problem \citep{Hyafil1976}. Historically, tree ensembles have been trained with two methods. The first method relies on greedy heuristics with bagging/boosting: where individual trees are trained with CART on bootstrapped samples of the data e.g., random forests (RF) \citep{Breiman2001} and its variants \citep{Geurts2006, Athey2019}; or sequentially/adaptively trained with gradient boosting: 
Gradient Boosting Decision Trees \citep{Hastie2009} and efficient variants~\citep{xgboost2016, lightgbm2017, catboost2018, Schapire2012}. Despite the success of ensemble methods, interesting challenges remain: 
(i) RF tend to under-perform gradient boosting methods such as XGBoost \citep{xgboost2016}. (ii) The tree ensembles are typically very large, making them a complex and hardly interpretable decision structure. Recent work by \cite{Carreira-Perpinan2018, Zharmagambetov2020} improve RF with local search methods via alternating minimization. However, their implementation is not open-source. (iii) Open-source APIs for gradient boosting are limited in terms of flexibility. They lack support for multi-task learning, handling missing responses or catering to customized loss functions. Modifying these APIs require significant effort and research
for custom applications.

The alternative approach for tree ensemble learning extends probabilistic/differentiable trees and performs end-to-end learning \citep{Kontschieder2015, Hazimeh2020}. These works build upon the idea of hierarchical mixture of experts introduced by \citep{Jordan1994} and further developed by \citep{Irsoy2012, Tanno2019, Frosst2017} for greedy construction of trees. 
Some of these works \citep{Kontschieder2015, Hazimeh2020} propose using differentiable trees as an output layer in a cascaded neural network for combining 
feature representation learning along with tree ensemble learning for classification. 
We focus on learning tree (ensembles) with hyperplane splits and constant leaf nodes--this allows us to expand the scope of trees to flexible loss functions, and develop specialized implementations that can be more efficient. 
One might argue that probabilistic trees are harder to interpret and suffer from slower inference as a sample must follow each root-leaf path, lacking {\emph{conditional computation}} present in classical decision trees. However, \cite{Hazimeh2020} proposed a principled way to get conditional inference in probabilistic trees by introducing a new activation function; this allows for routing samples through small parts of the tree similar to classical decision trees. We refer the reader to Section \ref{sec:expts-studying-single-tree} for a study on a single tree and highlight that the a soft tree with hyperplane splits and conditional inference has similar interpretability as that of a classical tree with hyperplane splits --- see Figure \ref{fig:single-tree}. Additionally, a soft tree can lead to smaller optimal 
depths---see Supplemental Section \ref{supp-sec:studying-single-tree}. 


End-to-end learning with differentiable tree ensembles appears to have several advantages. (i) They are easy to setup with public deep learning API e.g., Tensorflow \citep{tensorflow2015}, PyTorch \citep{pytorch2019}. We demonstrate that with a careful implementation, the tree ensembles can perform efficient GPU-training --- this is not possible with earlier toolkits e.g., TEL \citep{Hazimeh2020}. (ii) Tensorflow-Probability library \citep{tensorflowprob2017} offers huge flexibility in modeling. For instance, they allow for modeling mixture likelihoods, which can cater to zero-inflated data applications. This makes it possible to combine differentiable tree ensembles with customized losses with minimal effort. (iii) Deep learning APIs e.g., Tensorflow offer great support for handling multi-task loss objectives without a need for specialized algorithm. This makes it convenient to perform multi-task learning with differentiable tree ensembles. (iv) Differentiable trees can lead to more expressive and compact ensembles \citep{Hazimeh2020}. This can have important implications for interpretability, latency and storage requirements during inference.

\section{Optimizing Tree Ensembles}
\label{sec:optimization-framework-with-tree-ensembles}
We assume a supervised multi-task learning setting, with input space $\C{X} \subseteq \R^p$ and output space $\C{Y} \subseteq \R^k$. We learn a mapping $f: \R^p \rightarrow \R^k$, from input space $\mathcal{X}$ to output space $\mathcal{Y}$, where we parameterize function $f$ with a differentiable tree ensemble. We consider a general optimization framework where the learning objective is to minimize any differentiable function $g: \R^p \times \R^k \rightarrow \R$. The framework can accommodate different loss functions arising in different applications and perform end-to-end learning with tree ensembles. We first briefly review differentiable tree ensembles in section \ref{sec:background}. We later present a careful tensor-based implementation of the tree ensemble in section \ref{sec:efficient-tensor-formulation} that makes it possible to perform efficient training on both CPUs and GPUs. Next, we outline the need for flexible loss modeling and present some examples that our tree ensemble learning framework supports for learning compact tree ensembles in Sections \ref{sec:flexible-loss-functions} and \ref{sec:multi-task}. Finally, we present a metastudy on a large collection of real-world datasets to validate our framework.       


\paragraph{Notation}
We summarize our notation used in the paper. For an integer $n\ge1$, let $[n]:=\{1,2,....,n\}$. 
We let $1_m$ denote the vector in $\R^m$ with all coordinates being $1$. 
For matrix $\B B = ((B_{ij})) \in \R^{M\times N}$, let the $j$-th column be denoted by $\B B_{j}:= [B_{1j}, B_{2j}, ..., B_{Mj}]^T \in \R^M$ for $j \in [N]$. A dot product between two vectors $\B u, \B v \in R^m$ is denoted as $\B u \cdot \B v$. A dot product between a matrix $\B U \in R^{m,n}$ and a vector $\B v \in \R^m$ is denoted as $\B U \cdot \B v = \B U^T v \in \R^{n}$. A dot product between a tensor $\bm{\C{U}} \in \R^{p,m,n}$ and a vector $\B v \in \R^{m}$ is denoted as $\bm{\C{U}}\cdot \B v = \bm{\C{U}}^T \B v \in \R^{p,n}$ where the transpose operation of a tensor $\bm{\C{U}}^T \in \R^{p,n,m}$ permutes the last two dimensions of the tensor.

\subsection{Preliminaries and Setup}
\label{sec:background}
We learn an ensemble of $m$ differentiable trees. Let $\B f^{j}$ be the $j$th tree in the ensemble. For easier exposition, we consider a single-task regression or classification setting---see Section \ref{sec:multi-task} for an extension to the  multi-task setting. In a regression setting $k=1$, while in multi-class classification setting $k=C$, where $C$ is the number of classes. For an input feature-vector $\B x \in \R^p$, we learn an additive model with the output being sum over outputs of all the trees:
\begin{align}
    \B f(\B x) = \sum_{j=1}^m \B f^{j}(\B x).
    \label{eq:additive-trees}
\end{align}
The output, $\B f(\B x)$, is a vector in $\R^k$ containing raw predictions. For multiclass classification, mapping from raw predictions to $\C Y$ is done by applying a softmax function on the vector $\B f(\B x)$ and returning the class with the highest probability.
Next, we introduce the key building block of the approach: the differentiable decision tree.

\paragraph{Differentiable decision trees for modelling $\B f^{j}$}
Classical decision trees perform hard sample routing, i.e., a sample is routed to exactly one child at every splitting node.
Hard sample routing introduces discontinuities in the loss function, making trees unamenable to continuous optimization. Therefore, trees are usually built in a greedy fashion. 
In this section, we first introduce a single soft tree proposed by \cite{Jordan1994}, which is utilized in \cite{Irsoy2012, Frosst2017, Blanquero2021} and extended to soft tree ensembles in \cite{Kontschieder2015, Hehn2019, Hazimeh2020}.
A soft tree is a variant of a decision tree that performs soft routing, where every internal node can route the sample to the left and right simultaneously, with different proportions. 
This routing mechanism makes soft trees differentiable, so learning can be done using gradient-based methods. Notably, \cite{Hazimeh2020} introduced a new activation function for soft trees that allowed for conditional computation while preserving differentiability.

Let us fix some $j \in [m]$ and consider a single tree $\B f^j$ in the additive model (\ref{eq:additive-trees}). 
Recall that $\B f^j$ takes an input sample and returns an output vector (logit), i.e., $\B f^j: X \in \R^p \rightarrow \R^k$. 
Moreover, we assume that $\B f^j$ is a perfect binary tree with depth $d$. 
We use the sets $\C I^j$ and $\C L^j$ to denote the internal (split) nodes and the leaves of the tree, respectively. 
For any node $i \in \C{I}^j \cup \C{L}^j$, we define $A^j(i)$ as its set of ancestors and use the notation {$\B x \rightarrow i$} for the event that a sample $\B x \in \R^p$ reaches $i$.

\paragraph{Routing} Internal (split) nodes in a differentiable tree perform soft routing, where a sample is routed left and right with different proportions. This soft routing can be viewed as a probabilistic model. Although the sample routing is formulated with a probabilistic model, the final prediction of the tree $\B f$ is a deterministic function as it assumes an expectation over the leaf predictions. 
Classical decision trees are modeled with either axis-aligned splits \citep{Breiman1984, Quinlan1993} or hyperplane (a.k.a. oblique) splits \citep {Murthy1994}. Soft trees are based on hyperplane splits, where the routing decisions rely on a linear combination of the features. 
Particularly, each internal node $i \in \C{I}^j$ is associated with a trainable weight vector $\B{w}_i^j \in \R^p$ that defines the node's hyperplane split. 
Let $S: R \rightarrow [0, 1]$ be an activation function.
Given a sample $\B x \in \R^p$, the probability that internal node $i$ routes $\B x$ to the left is defined by $S(\B{w}_i^j \cdot \B x)$.
Now we discuss how to model the probability that $\B x$ reaches a certain leaf $l$.
Let $[l \shortarrow{5} i]$ (resp. [$i \shortarrow{7} l$]) denote the event that leaf $l$ belongs to the left (resp. right) subtree of node $i \in \C{I}^j$.
Assuming that the routing decision made at each internal node in the tree is independent of the other nodes, the probability that $\B x$ reaches $l$ is given by: 
\begin{align}
    P^j(\{x \rightarrow l\}) = \prod_{i \in A(l)} r_{i,l}^j(\B x),
    \label{eq:probability}
\end{align}
where $r_{i,l}^j(\B x)$ is the probability of node $i$ routing $\B x$ towards the subtree containing leaf $l$, i.e., $r_{i,l}^j(x) := S(\B{w}_i^j \cdot \B x) 1[l \shortarrow{5} i] \odot (1 - S(\B{w}_i^j \cdot \B x))1[i \shortarrow{7} l]$. Next, we define how the root-to-leaf probabilities in (\ref{eq:probability}) can be used to make the final prediction of the tree.

\paragraph{Prediction} As with classical decision trees, we assume that each leaf stores a weight vector $\B{o}_l^j \in R^k$ (learned during training).
Note that, during a forward pass, $\B{o}_l^j$ is a constant vector, meaning that it is not a function of the input sample(s).
For a sample $\B{x} \in \R^p$, we define the prediction of the tree as the expected value of the leaf outputs, i.e., 
\begin{align}
    \B f^j(\B x) = \sum_{l \in L} P^j(\{\B x \rightarrow l\})\B o_l^j.
    \label{eq:expectation}
\end{align}

\paragraph{Activation Function} In soft routing, the internal nodes use an activation function $S$ in order to compute the routing probabilities.
The logistic (a.k.a. sigmoid) function is the common choice for $S$ in the literature on soft trees (see \cite{Jordan1994, Kontschieder2015, Frosst2017, Tanno2019, Hehn2019}).
While the logistic function can output arbitrarily small values, it cannot output an exact zero.
This implies that any sample $x$ will reach every node in the tree with a positive probability (as evident from (\ref{eq:probability})).
Thus, computing the output of the tree in (\ref{eq:expectation}) will require computation over every node in the tree, an operation which is exponential in tree depth. \cite{Hazimeh2020} proposed a smooth-step activation function, which can output exact zeros and ones, thus allowing for true conditional computation. Furthermore, the smooth-step function allows efficient forward and backward propagation algorithms.

\subsection{Efficient Tensor Formulation}
\label{sec:efficient-tensor-formulation}
\begin{figure}
    \centering
    \includegraphics[width=0.6\columnwidth]{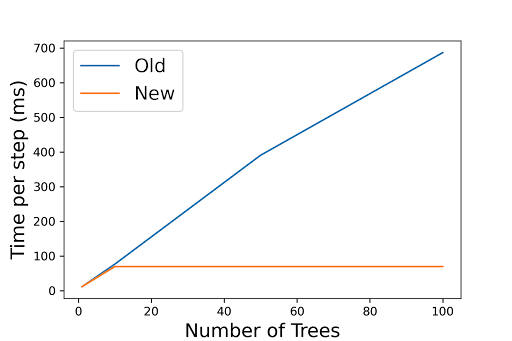}
    \caption{Timing comparison on CPU training with tensor-based implementation. GPU training with tensor-based implementation leads to an additional 40\% improvement over CPU training with tensor-based implementation.
    }
    \label{fig:timing-comparison-implementation}
\end{figure}

Current differentiable tree ensemble proposals and  toolkits, for example deep neural decision forests\footnote{\url{https://keras.io/examples/structured_data/deep_neural_decision_forests/}} \citep{Kontschieder2015} and TEL \citep{Hazimeh2020} model trees individually. This leads to slow CPU-training times and makes these implementations hard to vectorize for fast GPU training. In fact, TEL \citep{Hazimeh2020} doesn't support GPU training. We propose a tensor-based formulation of a tree ensemble that parallelizes routing decisions in nodes across the trees in the ensemble. This can lead to 10x faster CPU training times if the ensemble sizes are large e.g., $100$. Additionally, the tensor-based formulation is GPU-friendly, which provides an additional $40\%$ faster training times. See Figure \ref{fig:timing-comparison-implementation} for a timing comparison on CPU training without/with tensor formulation. Next, we outline the tensor-based formulation.

We propose to model the internal nodes in the trees across the ensemble jointly as a ``supernodes''.   
In particular, an internal node $i \in \C{I}^j$ at depth $d$ in all trees can be condensed together into a supernode $i \in \C{I}$.
We define a learnable weight matrix $\B W_i \in \R^{p,m}$, where each $j$-th column of the weight matrix contains the learnable weight vector $\B w_i^j$ of the original j-th tree in the ensemble. Similarly, the leaf nodes are defined to store a learnable weight matrix $\B O_l \in \R^{m,k}$, where each $j$-th row contains the learnable weight vector $\B o_l^j$ in the original $j$-th tree in the ensemble. The prediction of the tree with supernodes can be written as 
\begin{align}
\small
    \B f(\B x)= \left(\sum_{l \in L} \B O_l \odot \prod_{i\in A(l)}\B R_{i,l} \right)  \cdot \B{1}_m  
    \label{eq:multi-task-tree-prediction}
\end{align}
where $\odot$ denotes the element-wise product, $\B R_{i,l} = S(\B W_i \cdot \B x)1[l \shortarrow{5} i] \odot (1-S(\B W_i \cdot \B x))1[i \shortarrow{7} l] \in \R^{m,1}$ and the activation function $S$ is applied element-wise. This formulation of tree ensembles via supernodes allows for sharing of information across tasks via tensor formulation in multi-task learning ---  see Section \ref{sec:multi-task} for more details.   

\subsection{Toolkit} Our tree ensemble learning toolkit is built in Tensorflow (TF) 2.0 and integrates with Tensorflow-Probability. The toolkit allows the user to  write a custom loss function, and TF provides automatic differentiation. Popular packages, such as XGBoost, require users to provide first/second order derivatives. In addition to writing a custom loss, the user can select from a wide range of predefined loss and likelihood functions from Tensorflow-Probability. 
By relying on TF in the backend, our toolkit can easily exploit distributed computing. It can also run on multiple CPUs or GPUs, and on multiple  platforms, including mobile platforms.
Checkpointing for fault tolerance, incremental training and warm restarts are also supported. The toolkit will be open sourced if the paper is accepted.

\section{Flexible loss functions}
\label{sec:flexible-loss-functions}
Our framework can handle any differentiable loss function. Such flexibility is important as various applications require flexibility in loss functions beyond what is provided by current tree ensemble learning APIs. Our framework is built on Tensorflow, which gives us the capability to perform gradient-based optimization on scale. This coupled with our efficient differentiable tree ensemble formulation gives a powerful toolkit to seamlessly experiment with different loss functions and select what is suitable for the intended application. A few examples of flexible distributions that our toolkit supports --- due to compatibility with Tensorflow-Probability --- are normal, Poisson, gamma, exponential, mixture distributions e.g., zero-inflation models \citep{Lee2020}, and compound distributions e.g., negative binomial \citep{Yirga2020}. Other loss functions such as robustness to outliers \citep{Barron2019} can also be handled by our tree ensemble learning toolkit. To demonstrate the flexibility of our framework, we deeply investigate two specific examples: zero-inflated Poisson and negative binomial regression. These cannot be handled by the popular gradient boosting APIs such as XGBoost \citep{xgboost2016}, LightGBM \citep{lightgbm2017}. 

\paragraph{Zero-inflated Poisson Regression}
Zero-inflation occurs in many applications, e.g.,  understanding alcohol and drug abuse in young adults \cite{Jacobs2021}, characterizing undercoverage and overcoverage to gauge the on-going quality of the census frames \citep{Young2017}, studying popularity of news items on different social media platforms \cite{Moniz2018}, financial services applications \cite{Lee2020} etc.
Despite the prevalence of these applications, there has been limited work on building decision tree-based approaches for zero-inflated data perhaps due to a lack of support 
public APIs. Therefore, practitioners either resort to Poisson regression with trees or simpler linear models to handle zero-inflated responses. A Poisson model can lead to sub-optimal performance due to the limiting equidispersion constraint (mean equals the variance). Others take a two-stage approach \citep{Cameron2013}, where a classification model distinguishes the zero and non-zero and a second model is used to model the non-zero responses. This can be sup-optimal as errors in the first model can deteriorate the performance of the second model. We employ a more well-grounded approach by formulating the joint mixture model, where one part of the model tries to learn the mixture proportions (zero vs non-zero) and the other part models the actual non-zero responses. Such a mixture model permits a differentiable loss function when both components of the model are parameterized with differentiable tree ensembles and can be optimized with gradient descent method in an end-to-end fashion without the need for a custom solver. We provide an extensive study with our framework on small to large-scale real world zero-inflated datasets and demonstrate that such flexibility in distribution modeling can lead to significantly more compact and expressive tree ensembles. This has large implications for faster inference, storage requirements and interpretability.

We briefly review Poisson regression and then dive into zero-inflated Poisson models. Poisson regression stems from the generalized linear model (GLM) framework for modeling a response variable in the exponential family of distributions.
In general, GLM uses a link function to provide the relationship between the linear predictors, $\B x$ and the conditional mean of the density function:
\begin{align}
 g[\E(\ry|\B x)] = \B \beta \cdot \B x ,
\end{align}
where $\B \beta$ are parameters and $g(\cdot)$ is the link function. When responses $y_n$ (for $n \in [N])$, are independent and identically distributed (i.i.d.) and follow the Poisson distribution conditional on $\B x_n$'s, we use $log(\cdot)$ as the link function and call the model a Poisson regression model: $log(\mu_n|\B x_n) = \B \beta \cdot \B x_n$. We consider more general parameterizations with tree ensembles as given by
\begin{align}
    log(\mu_n|\B x_n) = f(\B x_n; \bm{\C W}, \bm{\C O}).
\end{align}
where $f$ is parameterized with a tree ensemble as in (\ref{eq:additive-trees}) and $\bm{\C W}, \bm{\C O}$ are the learnable parameters in the supernodes and the leaves of the tree ensemble. When a count data has excess zeros, the equi-dispersion assumption of the Poisson is violated. The Poisson model is not an appropriate model for this situation anymore. \cite{Lambert1992} proposed zero-inflated-Poisson (ZIP) models that address the mixture of excess zeros and Poisson count process.
The mixture is indicated by the latent binary variable $d_n$ using a logit model and the density for the Poisson count given by the log-linear model. Thus,
\begin{align}
    y_n = \begin{cases}
    0, &\text{if}~~d_n = 0 \\
    y_n^*, &\text{if}~~d_n = 1
    \end{cases}
\end{align}
where the latent indicator $d_n \sim Bernoulli(\pi_n)$ with $\pi_n = P(d_n = 1)$ and $y_n^* \sim Poisson(\mu_n)$.
The mixture yields the marginal probability mass function of the observed $y_n$ given as:
\begin{align}
    ZIP(y_n | \mu_n, \pi_n) = \begin{cases}
    (1 - \pi_n) + \pi_n e^{-\mu_n}, &\text{if}~~y_n=0 \\
    \pi_n e^{-\mu_n} \mu_n^{y_n}/y_n!, &\text{if}~~y_n=1,2,\cdots    
    \end{cases}
\end{align}
where $\mu_n$ and $\pi_n$ are modeled by
\begin{align}
    log\left(\frac{\pi_n}{1-\pi_n}|\B x_n\right) &=  f(\B x_n; \bm{\C Z}, \bm{\C U}) \\
    log(\mu_n|\B x_n) &= f(\B x_n; \bm{\C W}, \bm{\C O}).
\end{align}
where $\bm{\C Z}, \bm{\C U}$ are the learnable parameters in the splitting internal supernodes and the leaves of the tree ensemble for the logit model for $\pi_n$ and $\bm{\C W}, \bm{\C O}$ are the learnable parameters in the supernodes and the leaves of the tree ensemble for the log-tree model for $\mu_n$ respectively.
The likelihood function for this ZIP model is given by
\begin{align}
    L(y_n, f(\B x_n)) = \prod_{y_n = 0} (1 - \pi_n) + \pi_n e^{-\mu_n} \prod_{y_n > 0} \pi_n e^{-\mu_n} \mu_n^{y_n}/y_n!
\end{align}
where $\lambda_n = e^{f(\B x_n; \bm{\C W}, \bm{\C O})}$ and $\pi_n = e^{f(\B x_n, \bm{\C Z}; \bm{\C U})}/ (1+e^{f(\B x_n; \bm{\C Z}, \bm{\C U})})$.Such a model can be overparameterized and we observed that sharing the learnable parameters $\bm{\C Z}=\bm{\C W}$ in the splitting internal supernodes across the log-mean and logit models can lead to better out-of-sample performance --- see Section \ref{sec:experiments} for a thorough evaluation on real-world datasets. 

\paragraph{Negative Binomial Regression}
An alternative distribution to zero-inflation modeling that can cater to over-dispersion in the responses is  Negative Binomial (NB) distribution. A negative binomial distribution for a random variable $\ry$ with a non-negative mean $\mu \in \R_{+}$ and dispersion parameters $\phi \in \R_{+}$ is given by:
\begin{align}
    NB (\ry | \mu, \phi) = \left(\frac{\ry + \phi - 1}{\ry}\right)\left(\frac{\mu}{\mu + \phi}\right)^{\ry} \left(\frac{\phi}{\mu + \phi}\right)^{\phi}
\end{align}
The mean and variance of a random variable $\ry \sim NB(\ry | \mu, \phi)$ are $\E[\ry] = \mu$ and $\Var[\ry] = \mu + \mu^2/\phi$. Recall that Poisson($\mu$) has variance $\mu$, so  
$\mu^2 / \phi > 0$ is the additional variance of the negative binomial above that of the Poisson with mean $\mu$. So the inverse of parameter $\phi$ controls the overdispersion, scaled by the square of the mean, $\mu^2$.

When the responses $y_n$ (for $n \in [N]$) are i.i.d, and follow NB distribution conditioned on $\B x_n$'s, we can use the $log(.)$ as a link function to parameterize the log-mean and log-dispersion as linear functions of the covariates $\B x_n$. In our parameterization with Tree Ensembles, we model them as given by:
\begin{align}
    log(\mu_n | \B x_n) &= f(\B x_n; \bm{\C W}, \bm{\C O}) \\
    log\left(\phi_n |\B x_n \right) &= f(\B x_n; \bm{\C Z}, \bm{\C U}).
\end{align}
where $\bm{\C Z}, \bm{\C U}$ are the learnable parameters in the supernodes and the leaves of the tree ensemble for the log-mean and $\bm{\C W}, \bm{\C O}$ are the learnable parameters in the supernodes and the leaves of the tree ensemble for the log-dispersion model for $\phi_n$ respectively. Such a model can be overparameterized and we observed that sharing the learnable parameters $\bm{\C Z}=\bm{\C W}$ in the splitting internal supernodes across the log-mean and log-dispersion models can lead to better out-of-sample performance. See Section \ref{sec:expts-lm} for empirical validation on a large-scale dataset.

\section{Multi-task Learning with Tree Ensembles}
\label{sec:multi-task}
Multi-task Learning (MTL) aims to learn multiple tasks simultaneously by using a shared model. Unlike single task learning, MTL can achieve better generalization performance through exploiting task relationships \citep{Caruana1997, Chapelle2010}.
One key problem in MTL is how to share model parameters between tasks \citep{Ruder2017}.
For instance, sharing parameters between unrelated tasks can potentially degrade performance. MTL approaches for classical decision trees approaches e.g., RF \citep{Linusson2013}, GRF \citep{Athey2019} have shared weights at the splitting nodes across the tasks.
Only the leaf weights are task specific. 
However this can be limiting in terms of performance, despite easier interpretability associated with the same split nodes across tasks.

To perform flexible multi-task learning, we extend our formulation in Section \ref{sec:efficient-tensor-formulation} by using task-specific nodes in the tree ensemble.
We consider $T$ tasks.
For easier exposition, we consider tasks of the same kind: multilabel classification or multi-task regression. For multilabel classification, each task is assumed to have same number of classes (with $k=C$) for easier exposition --- our framework can handle multilabel settings with different number of classes per task. Similarly, for regression settings, $k=1$. 
For multi-task zero-inflated Poisson or negative binomial regression, when two model components need to be estimated, we set $k=2$ to predict log-mean and logit components for zero-inflated Poisson and log-mean and log-dispersion components for negative binomial.

We define a trainable weight tensor $\bm{\C W}_i \in \R^{T,p,m}$ for supernode $i \in \C I$, where each $t$-th slice of the tensor $\bm{\C W}_i [t,:,:]$ denotes the trainable weight matrix associated with task $t$. The prediction in this case is given by  
\begin{align}
    \B f(\B x)= \biggl(\sum_{l \in L} \bm{\C O}_l \odot \prod_{i\in A(l)} \bm{\C{R}}_{i,l} \biggr)  \cdot \B{1}_{m}  \label{eq:multi-task-tree-prediction-with-task-specific-supernodes}
\end{align}
where $\bm{\C{O}}_l \in \R^{T,m,k}$ denotes the trainable leaf tensor in leaf $l$, $\bm{\C{R}}_{i,l}=S(\bm{\C W}_i \cdot \B x)1[l \shortarrow{5} i] \odot (1-S(\bm{\C W}_i \cdot \B x))1[i \shortarrow{7} l] \in \R^{T,m,1}$.

In order to share information across the tasks, our framework imposes closeness penalty on the hyperplanes $\bm{\C{W}}_i$ in the supernodes across the tasks. This results in an optimization formulation:
\begin{align}
    \min_{\bm{\C {W}}, \bm{\C O}} \sum_{t \in T} \sum_{\B x, y_t} g_t( y_t, f_t(\B x)) + \lambda \sum_{s<t, t \in T}\norm{\bm{\C{W}}_{:,s,:,:} - \bm{\C{W}}_{:,t,:,:}}^2,
    \label{eq:optimization}
\end{align}
where $\bm{\C W} \in \R^{\C I, T, m, p}$ denotes all the weights in all the supernodes, $\bm{\C O} \in \R^{\C L, m, k}$ denotes all the weights in the leaves, and $\lambda \in [0,\infty)$ is a non-negative regularization penalty that controls how close the weights across the tasks are. The model behaves as single-task learning when $\lambda = 0$ and complete sharing of information in the splitting nodes when $\lambda \rightarrow \infty$. Note that when $\lambda \rightarrow \infty$, the weights across the tasks in each of the internal supernodes become the same. This case can be separately handled more efficiently by using the function definition in (\ref{eq:multi-task-tree-prediction}) for $\B f(\B x)$ without any closeness regularization in (\ref{eq:optimization}). 
Our model can control the level of sharing across the tasks by controlling $\lambda$. In practice, we tune over $\lambda \in [1e-5, 10]$ and select the optimal based on the validation set. This penalty assumes that the hyperplanes across the tasks should be equally close as we go down the depth of the trees. However this assumption maybe less accurate as we go down the tree. Empirically, we found that decaying $\lambda$ exponentially as $\lambda/2^{d}$ with depth $d$ of the supernodes in the ensemble can achieve better test performance.

\section{Experiments}
\label{sec:experiments}
We study the performance of differentiable tree ensembles in various settings and compare against the relevant state-of-the-art baselines for each setting. The different settings can be summarized as follows: (i) Comparison of a single soft tree with state-of-the-art classical tree method in terms of test performance and depth. We include both axis-aligned and oblique classical tree in our comparisons. (ii) Flexible zero-inflation models with tree ensembles. We compare against Poisson regression with tree ensembles and gradient boosting decision trees (GBDT). We consider test Poisson deviance and tree ensemble compactness for model evaluation (iii) We evaluate our proposed multi-task tree ensembles and compare them against multioutput RF, multioutput GRF and single-task GBDT. We consider both fully observed and partially observed responses across tasks. (iv) We also validate our tree ensemble methods with flexible loss functions (zero-inflated Poisson and negative binomial regression) on a large-scale multi-task proprietary dataset. 


\paragraph{Model Implementation} Differentiable tree ensembles are implemented in TensorFlow 2.0 using Keras interface.

\paragraph{Datasets} We use 27 open-source regression datasets from various domains (e.g., social media platforms, human behavior, finance). 9 are from Mulan \citep{Xioufis2016}, 2 are from UCI data repository \citep{Dua2019}, 12 are from Delve database \citep{Akujuobi2017} and the 5 remaining are SARCOS \citep{Vijayakumar2000}, Youth Risk Behavior Survey \citep{Jacobs2021}, Block-Group level Census data \citep{Bureau2021}, and a financial services loss dataset from Kaggle\footnote{\url{https://bit.ly/3swGnTo}}. We also validate our framework on a proprietary multi-task data with millions of samples from a multi-national financial services company.


\subsection{Studying a single tree}
\label{sec:expts-studying-single-tree}
In this section, we compare performance and model compactness of a single tree on 12 regression datasets from Delve database: abalone, pumadyn-family, 
comp-activ (cpu, cpuSmall) and concrete.

\paragraph{Competing Methods and Implementation} We focus on two baselines from classical tree literature: CART \citep{Breiman1984} and Tree Alternating Optimization (TAO) method proposed by \cite{Carreira-Perpinan2018}. The authors in \cite{Zharmagambetov2020} performed an extensive comparison of various single tree learners and demonstrated TAO to be the best performer. Hence, we include both axis-aligned and oblique decision tree versions of TAO in our comparisons. Given that the authors in \cite{Carreira-Perpinan2018, Zharmagambetov2020} do not provide an open-source implementation for TAO, we implemented our own version of TAO. For a fair comparison, we use binary decision trees for both axis-aligned and oblique versions of TAO. For more details about our implementation, see Supplemental Section \ref{supp-sec:studying-single-tree}.

\begin{table}[!tb]
\centering
\caption{Test mean squared error performance of a \textit{single} axis aligned and oblique decision tree on various regression datasets.}
\label{tab:single-tree-mse}
\small
\setlength{\tabcolsep}{17.0pt}
\resizebox{0.9\textwidth}{!}{\begin{tabular}{|l|cc|cc|}
\hline
 & \multicolumn{2}{c|}{\textbf{Axis-Aligned}} & \multicolumn{2}{c|}{\textbf{Oblique}} \\ \hline
\textbf{Data} & \multicolumn{1}{c|}{\textbf{CART}} & \textbf{TAO} & \multicolumn{1}{c|}{\textbf{TAO}} & \textbf{Soft Tree} \\ \hline
abalone & \multicolumn{1}{c|}{7.901E-03} & 8.014E-03 & \multicolumn{1}{c|}{7.205E-03} & \textbf{6.092E-03} \\ \hline
pumadyn-32nh & \multicolumn{1}{c|}{9.776E-03} & 9.510E-03 & \multicolumn{1}{c|}{1.146E-02} & \textbf{8.645E-03} \\ \hline
pumadyn-32nm & \multicolumn{1}{c|}{2.932E-03} & 2.741E-03 & \multicolumn{1}{c|}{4.942E-03} & \textbf{1.280E-03} \\ \hline
pumadyn-32fh & \multicolumn{1}{c|}{1.324E-02} & 1.307E-02 & \multicolumn{1}{c|}{1.370E-02} & \textbf{1.166E-02} \\ \hline
pumadyn-32fm & \multicolumn{1}{c|}{2.246E-03} & 2.177E-03 & \multicolumn{1}{c|}{3.566E-03} & \textbf{1.602E-03} \\ \hline
pumadyn-8nh & \multicolumn{1}{c|}{1.995E-02} & 1.983E-02 & \multicolumn{1}{c|}{2.027E-02} & \textbf{1.670E-02} \\ \hline
pumadyn-8nm & \multicolumn{1}{c|}{4.878E-03} & 4.584E-03 & \multicolumn{1}{c|}{4.206E-03} & \textbf{2.352E-03} \\ \hline
pumadyn-8fh & \multicolumn{1}{c|}{2.182E-02} & 2.200E-02 & \multicolumn{1}{c|}{2.159E-02} & \textbf{2.048E-02} \\ \hline
pumadyn-8fm & \multicolumn{1}{c|}{4.398E-03} & 4.347E-03 & \multicolumn{1}{c|}{4.074E-03} & \textbf{3.543E-03} \\ \hline
cpu & \multicolumn{1}{c|}{9.655E-04} & 1.475E-03 & \multicolumn{1}{c|}{1.312E-03} & \textbf{9.159E-04} \\ \hline
cpuSmall & \multicolumn{1}{c|}{1.450E-03} & 2.319E-03 & \multicolumn{1}{c|}{1.171E-03} & \textbf{9.159E-04} \\ \hline
concrete & \multicolumn{1}{c|}{7.834E-03} & 6.619E-03 & \multicolumn{1}{c|}{1.166E-02} & \textbf{4.139E-03} \\ \hline
\end{tabular}}
\end{table}

\begin{figure}
    \centering
    \begin{subfigure}[b]{0.3\columnwidth}  
        \centering 
        \includegraphics[width=\textwidth]{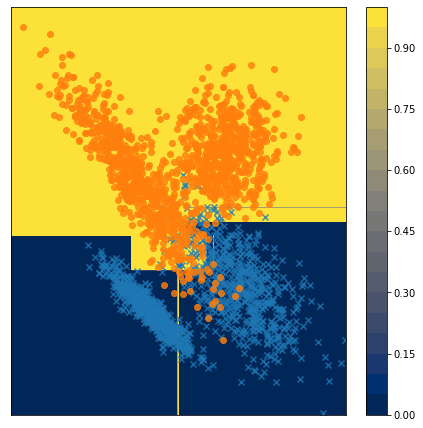}
    \end{subfigure}
    \begin{subfigure}[b]{0.3\columnwidth}   
        \centering 
        \includegraphics[width=\columnwidth]{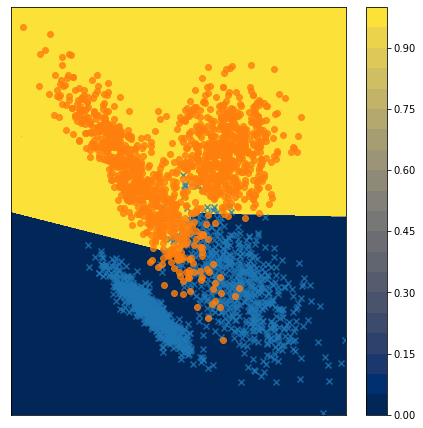}
    \end{subfigure}
    \begin{subfigure}[b]{0.3\columnwidth}   
        \centering 
        \includegraphics[width=\columnwidth]{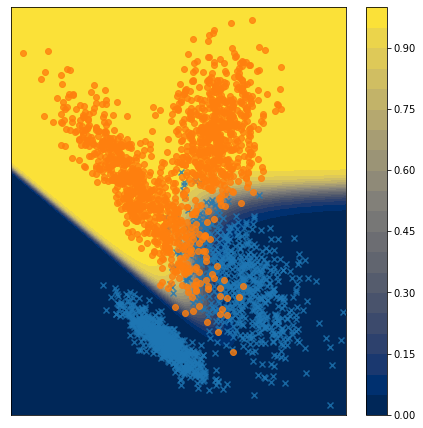}
    \end{subfigure}
    \caption[]
    {\small Classifier boundaries for CART [Left], TAO (oblique) [Middle] and Soft tree [Right] on a synthetic dataset with $N_{train}=N_{val}=N_{test}=2500$ generated using sklearn \citep{sklearn2013}. We tune for 50 trials over depths in the range $[2-4]$ for TAO (oblique) and soft trees and $[2-10]$ for CART. Optimal depths for CART, TAO, Soft tree are 5, 4 and 2 respectively. Test AUCs are 0.950, 0.957, and 0.994 respectively.} 
    \label{fig:single-tree}
\end{figure}

\paragraph{Results} We present the out-of-sample mean-squared-error performance and optimal depths in Tables \ref{tab:single-tree-mse} and \ref{tab:single-tree-depth} (in Supplemental Section \ref{supp-sec:studying-single-tree}) respectively. Notably, in all $12$ cases, soft tree outperforms all 3 baseline methods in terms of test performance. The soft tree finds a smaller optimal depth in majority cases in comparison with its classical counterpart i.e., oblique TAO tree --- See Table \ref{tab:single-tree-depth} in Supplemental Section \ref{supp-sec:studying-single-tree}. This may be due to the end-to-end learning in a soft tree, unlike TAO that performs local search.

\subsection{Zero-inflation}
\label{sec:expts-zero-inflation}
We consider a collection of real-world applications with zero-inflated data. The datasets include (i) yrbs: nationwide drug use behaviors of high school students as a function of demographics, e-cigarettes/mariyuana use etc.; (ii) news: popularity of news items on social media platforms \cite{Moniz2018} e.g., Facebook, Google+ as a function of topic and sentiments; (iii) census: number of people with zero, one, or two health insurances across all Census blocks in the ACS population as a function of housing and socio-economic demographics; (iv) fin-services-loss: financial services losses as a function of geodemographics, information on crime rate,  weather.

\paragraph{Competing methods} We consider Poisson regression with GBDT and differentiable tree ensembles. We also consider zero-inflation modeling with differentiable tree ensembles. We use GBDT from sklearn \cite{sklearn2013}. For additional details about the tuning experiments, please see Supplemental Section \ref{supp-sec:expts-zero-inflation}.

\begin{table}[!tb]
\centering
\small
\caption{Flexible modeling via zero-inflated Poisson for Soft Tree Ensembles leads to better out-of-sample performance.}
\label{tab:poisson-deviance-zip}
\setlength{\tabcolsep}{14.0pt}
\resizebox{0.9\textwidth}{!}{\begin{tabular}{|lrr|ccc|}
\hline
 &  &  &  \multicolumn{1}{c|}{\textbf{GBDT}} & \multicolumn{2}{c|}{\textbf{Soft Trees}} \\ \hline
\multicolumn{1}{|l|}{\textbf{Data}} & \multicolumn{1}{c|}{\textbf{N}} & \multicolumn{1}{c|}{\textbf{p}} &  \multicolumn{2}{c|}{\textbf{Poisson}} & \textbf{ZIP} \\ \hline
\multicolumn{1}{|l|}{yrbs-cocaine} & \multicolumn{1}{r|}{12172} & \multicolumn{1}{r|}{55} & 3.14E-02 & 3.00E-02 & \textbf{2.82E-02} \\ \cline{1-6} 
\multicolumn{1}{|l|}{yrbs-heroine} & \multicolumn{1}{r|}{12711} & \multicolumn{1}{r|}{55} & 1.81E-02 & 1.60E-02 & \textbf{1.54E-02} \\ \cline{1-6} 
\multicolumn{1}{|l|}{yrbs-meth} & \multicolumn{1}{r|}{12690} & \multicolumn{1}{r|}{55} & 2.38E-02 & 2.21E-02 & \textbf{2.09E-02} \\ \cline{1-6} 
\multicolumn{1}{|l|}{yrbs-lsd} & \multicolumn{1}{r|}{9564} & \multicolumn{1}{r|}{55} & 3.51E-02 &  3.59E-02 & \textbf{3.43E-02} \\ \hline
\multicolumn{1}{|l|}{news-facebook} & \multicolumn{1}{r|}{81637} & \multicolumn{1}{r|}{3} & 4.70E-03 & \textbf{4.68E-03} & \textbf{4.68E-03} \\ \cline{1-6} 
\multicolumn{1}{|l|}{news-google+} & \multicolumn{1}{r|}{87495} & \multicolumn{1}{r|}{3} & 5.97E-03 & \textbf{5.93E-03} & \multicolumn{1}{l|}{\textbf{5.93E-03}} \\ \cline{1-6} \hline
\multicolumn{1}{|l|}{census-health0} & \multicolumn{1}{r|}{220333} & \multicolumn{1}{r|}{64} & 1.51E-04 & \textbf{1.28E-04} & 1.32E-04 \\ \cline{1-6} 
\multicolumn{1}{|l|}{census-health1} & \multicolumn{1}{r|}{220333} & \multicolumn{1}{r|}{64} & \textbf{4.63E-04} & 5.11E-04 & 4.66E-04 \\ \cline{1-6} 
\multicolumn{1}{|l|}{census-health2+} & \multicolumn{1}{r|}{220333} & \multicolumn{1}{r|}{64} & 2.73E-03 & 3.06E-03 & \textbf{2.72E-03} \\ \hline
\multicolumn{1}{|l|}{fin-services-losses} & \multicolumn{1}{r|}{452061} & \multicolumn{1}{r|}{300} & \textbf{2.20E-03} & 2.28E-03 & \textbf{2.20E-03} \\ \hline
\multicolumn{1}{|l|}{\#wins} & - & - & 2 & 3 & 8 \\ \hline
\end{tabular}}
\end{table}

\begin{table}[!tb]
\centering
\caption{Flexible modeling via zero-inflated Poisson for Soft Tree Ensembles can lead to more compact tree ensembles, which can potentially lead to easier interpretability}
\small
\label{tab:model-sizes-zip}
\setlength{\tabcolsep}{15.0pt}
\resizebox{0.9\textwidth}{!}{\begin{tabular}{|l|ccc|ccc|}
\hline
 & \multicolumn{3}{c|}{\textbf{\#Trees}} & \multicolumn{3}{c|}{\textbf{Depth}} \\ \cline{2-7} 
 & \multicolumn{1}{l|}{\textbf{GBDT}} & \multicolumn{2}{l|}{\textbf{Soft Trees}} & \multicolumn{1}{l|}{\textbf{GBDT}} & \multicolumn{2}{l|}{\textbf{Soft Trees}} \\ \hline
\textbf{Data} & \multicolumn{2}{c|}{\textbf{Poisson}} & \textbf{ZIP} & \multicolumn{2}{c|}{\textbf{Poisson}} & \textbf{ZIP} \\ \hline
yrbs-cocaine & 575 & 77 & \textbf{13} & 4 & \textbf{2} & 3 \\ \hline
yrbs-heroine & 1425 & 83 & \textbf{4} & 4 & 4 & \textbf{2} \\ \hline
yrbs-meth & 1475 & 16 & \textbf{4} & 4 & \textbf{2} & 3 \\ \hline
yrbs-lsd & 1225 & \textbf{17} & 45 & 4 & 3 & \textbf{2} \\ \hline
news-facebook & 200 & \textbf{65} & 85 & 4 & 4 & 4 \\ \hline
news-google+ & 750 & 81 & \textbf{74} & 4 & 4 & 4 \\ \hline
census-health0 & 1275 & \textbf{10} & \textbf{10} & 4 & 3 & \textbf{2} \\ \hline
census-health1 & 1275 & \textbf{17} & \textbf{17} & 4 & 2 & \textbf{2} \\ \hline
census-health2+ & 1375 & 73 & \textbf{56} & 8 & 4 & \textbf{3} \\ \hline
fin-services-losses & 1225 & 32 & \textbf{4} & 4 & 3 & \textbf{2} \\ \hline
\#wins & 0 & 4 & \textbf{7} &  - & 5 & \textbf{8} \\ \hline
\end{tabular}}
\end{table}

\paragraph{Results} We present the out-of-sample Poisson deviance performance in Table \ref{tab:poisson-deviance-zip}. Notably, tree ensembles with zero-inflated loss function leads the chart. We also present the optimal selection of tree ensemble sizes and depths in Table \ref{tab:model-sizes-zip}. We can observe that zero-inflation modeling can lead to significant benefits in terms of model compression. Both tree ensemble sizes and depths can potentially be made smaller, which have implications for faster inferences, memory footprint and interpretability.

\begin{table}[!bt]
\centering
\small
\caption{Performance of RF, GRF and multi-task Differentiable Tree Ensembles on 11 multi-task regression datasets with fully observed responses across tasks.}
\label{tab:fully-observed-response}
\setlength{\tabcolsep}{25.0pt}
\resizebox{0.9\textwidth}{!}{\begin{tabular}{|l|c|ccc|}
\hline
 &  & \multicolumn{3}{c|}{\textbf{Multi-task}} \\ \hline
\textbf{Data} & \textbf{Task} & \multicolumn{1}{c|}{\textbf{RF}} & \multicolumn{1}{c|}{\textbf{GRF}} & \textbf{Soft Trees} \\ \hline
 & 1 & 2.242E-02 & 1.847E-02 & \textbf{5.383E-03}  \\
 & 2 & 2.498E-02 & 1.894E-02 & \textbf{7.217E-03} \\
 & 3 & 1.127E-02 & \textbf{9.625E-03} & 1.128E-02 \\
 & 4 & 1.574E-02 & 1.504E-02 & \textbf{1.403E-02} \\
 & 5 & 2.040E-02 & 1.905E-02 & \textbf{1.182E-02} \\
\multirow{-6}{*}{atp1d} & 6 & 1.571E-02 & \textbf{1.333E-02} & 1.527E-02 \\ \hline
 & 1 & 3.244E-03 & \textbf{2.691E-03} & 8.965E-03 \\
 & 2 & \textbf{3.914E-03} & 3.931E-03 & 4.456E-03 \\
 & 3 & 1.231E-02 & 1.078E-02 & \textbf{1.059E-02} \\
 & 4 & \textbf{5.459E-03} & 6.542E-03 & 6.001E-03 \\
 & 5 & \textbf{1.842E-03} & 1.922E-03 & 3.358E-03 \\
\multirow{-6}{*}{atp7d} & 6 & \textbf{4.042E-03} & 5.303E-03 & 5.033E-03 \\ \hline
 & 1 & 4.384E-02 & \textbf{3.151E-02} & 3.345E-02 \\
 & 2 & 1.077E-02 & 4.638E-03 & \textbf{3.828E-03} \\
\multirow{-3}{*}{sf1} & 3 & 4.252E-02 & \textbf{2.863E-02} & 2.993E-02 \\ \hline
 & 1 & 7.883E-03 & 8.807E-03 & \textbf{7.789E-03} \\
 & 2 & 3.247E-03 & 2.583E-03 & \textbf{2.206E-03} \\
\multirow{-3}{*}{sf2} & 3 & \textbf{1.262E-03} & 3.744E-03 & 3.595E-03 \\ \hline
 & 1 & 3.027E-02 & 3.008E-02 & \textbf{2.233E-02} \\
 & 2 & 1.483E-02 & 1.405E-02 & \textbf{1.015E-02} \\
\multirow{-3}{*}{jura} & 3 & 7.896E-03 & 7.586E-03 & \textbf{6.036E-03} \\ \hline
 & 1 & 2.473E-04 & \textbf{1.865E-04} & 2.063E-04 \\
\multirow{-2}{*}{enb} & 2 & 2.830E-03 & 3.305E-03 & \textbf{1.054E-03} \\ \hline
 & 1 & 1.732E-01 & 1.396E-01 & \textbf{1.001E-01} \\
 & 2 & 1.224E-01 & 9.827E-02 & \textbf{7.368E-02} \\
\multirow{-3}{*}{slump} & 3 & 3.878E-02 & 2.944E-02 & \textbf{5.149E-03} \\ \hline
 & 1 & 3.040E-03 & 2.530E-03 & \textbf{1.794E-03} \\
 & 2 & 3.397E-03 & 3.003E-03 & \textbf{2.226E-03} \\
 & 3 & 4.178E-03 & 3.611E-03 & \textbf{2.940E-03} \\
\multirow{-4}{*}{scm1d} & 4 & 3.991E-03 & 3.376E-03 & \textbf{2.150E-03} \\ \hline
 & 1 & 4.457E-03 & 3.650E-03 & \textbf{2.198E-03} \\
 & 2 & 4.766E-03 & 3.632E-03 & \textbf{2.410E-03} \\
 & 3 & 4.892E-03 & 3.506E-03 & \textbf{2.620E-03} \\
\multirow{-4}{*}{scm20d} & 4 & 5.573E-03 & 4.072E-03 & \textbf{2.632E-03} \\ \hline
 & 1 & 3.466E-03 & 2.558E-03 & \textbf{1.730E-03} \\
 & 2 & 4.636E-03 & 4.039E-03 &  \textbf{3.728E-03}  \\
\multirow{-3}{*}{bike} & 3 & 5.123E-03 & 4.303E-03 & \textbf{3.822E-03} \\ \hline
\multirow{-1}{*}{\# wins} & - & 5 & 6 & \textbf{26} \\ \hline
\end{tabular}}
\end{table}

\begin{table}[!tb]
\centering
\small
\caption{Tree ensemble sizes for soft trees, RF, and GRF.}
\label{tab:tree-ensemble-sizes}
\setlength{\tabcolsep}{7.0pt}
\resizebox{0.9\columnwidth}{!}{\begin{tabular}{|c|c|c|c|c|c|c|c|c|c|c|}
\hline
\multicolumn{1}{|l|}{\textbf{}} & \multicolumn{1}{l|}{\textbf{atp1d}} & \multicolumn{1}{l|}{\textbf{atp7d}} & \multicolumn{1}{l|}{\textbf{sf1}} & \multicolumn{1}{l|}{\textbf{sf2}} & \multicolumn{1}{l|}{\textbf{jura}} & \multicolumn{1}{l|}{\textbf{enb}} & \multicolumn{1}{l|}{\textbf{slump}} & \multicolumn{1}{l|}{\textbf{scm1d}} & \multicolumn{1}{l|}{\textbf{scm20d}} & \multicolumn{1}{l|}{\textbf{bike}} \\ \hline
\textbf{RF} & 100 & 125 & 500 & 225 & 150 & 100 & 825 & 175 & 100 & 100 \\ \hline
\textbf{GRF} & 1050 & 950 & 350 & 100 & 150 & 100 & 350 & 50 & 100 & 250 \\ \hline
\textbf{Ours} & 10 & 15 & 12 & 44 & 17 & 13 & 54 & 49 & 25 & 93 \\ \hline
\end{tabular}}
\end{table}

\subsection{Multi-task Regression}
\label{sec:expts-multi-task}
We compare performance and model compactness of our proposed regularized multi-task tree ensembles on 11 multi-task regression datasets from Mulan (atp1d, atp7d, sf1, sf2, jura, enb, slump, scm1d, scm20d), and UCI data repository (bike) and SARCOS dataset.

\paragraph{Competing Methods}
We focus on 4 tree ensemble baselines from literature: single-task soft tree ensembles, sklearn GBDT, sklearn multioutput RF \cite{sklearn2013} and r-grf package for GRF \cite{Athey2019}. We consider two multi-task settings: (i) All Fully observed responses for all tasks, (ii) Partially observed responses across tasks. In the former case, we compare against RF and GRF. In the latter case, we compare against single-task soft tree ensembles and GBDT. Note the open-source implementations for RF and GRF do not support partially observed responses for multi-task settings and GBDT does not have support for multi-task setting. We refer the reader to Supplemental Section \ref{supp-sec:expts-multi-task} for tuning experiments details.

\begin{table}[!htb]
\centering
\small
\caption{Performance of GBDT, single-task and multi-task Soft Tree Ensembles on 11 multi-task regression datasets with 50\% missing responses per task.}
\label{tab:partially-observed-response}
\setlength{\tabcolsep}{20.0pt}
\resizebox{0.9\textwidth}{!}{\begin{tabular}{|ll|ccc|}
\hline
 &  & \multicolumn{2}{c|}{\textbf{Single-Task}} & \textbf{Multi-Task} \\ \hline
\multicolumn{1}{|l|}{\textbf{Data}} & \textbf{Task} & \multicolumn{1}{c|}{\textbf{GBDT}} & \multicolumn{2}{c|}{\textbf{Soft Trees}} \\ \hline
\multicolumn{1}{|l|}{} & 1 & 1.469E-02 & 3.091E-02 & \textbf{1.295E-02} \\
\multicolumn{1}{|l|}{} & 2 & \textbf{7.698E-03} & 2.598E-02 & 1.137E-02  \\
\multicolumn{1}{|l|}{} & 3 & 2.172E-02 & 2.915E-02 & \textbf{1.807E-02} \\
\multicolumn{1}{|l|}{} & 4 & \textbf{5.905E-03} & \textbf{1.417E-02} &  9.434E-03 \\
\multicolumn{1}{|l|}{} & 5 & 2.421E-02 & 5.631E-02 & \textbf{2.105E-02}  \\
\multicolumn{1}{|l|}{\multirow{-6}{*}{atp1d}} & 6 & \textbf{5.646E-03} & 5.880E-02 & 2.724E-02 \\ \hline
\multicolumn{1}{|l|}{} & 1 & 5.562E-02 & 6.261E-02 & \textbf{3.601E-02}  \\
\multicolumn{1}{|l|}{} & 2 & 4.033E-02 & 2.216E-02 & \textbf{1.067E-02} \\
\multicolumn{1}{|l|}{} & 3 & 2.140E-02 & 4.989E-02 & \textbf{1.680E-02} \\
\multicolumn{1}{|l|}{} & 4 & 1.107E-02 & 2.089E-02 & \textbf{9.926E-03} \\
\multicolumn{1}{|l|}{} & 5 & 4.254E-02 & 6.476E-02 & \textbf{3.053E-02} \\
\multicolumn{1}{|l|}{\multirow{-6}{*}{atp7d}} & 6 & \textbf{4.195E-03} & 2.416E-02 & ,  2.199E-02 \\ \hline
\multicolumn{1}{|l|}{} & 1 & \textbf{2.674E-02} & 2.763E-02 & 2.732E-02 \\
\multicolumn{1}{|l|}{} & 2 & 5.825E-03 & 1.680E-02 & \textbf{5.435E-03} \\
\multicolumn{1}{|l|}{\multirow{-3}{*}{sf1}} & 3 & 3.030E-02 & 3.704E-02 & \textbf{2.964E-02} \\ \hline
\multicolumn{1}{|l|}{} & 1 & \textbf{1.003E-02} & 1.179E-02 & \textbf{1.009E-02} \\
\multicolumn{1}{|l|}{} & 2 & 1.123E-02 & 6.843E-03 & \textbf{2.809E-03} \\
\multicolumn{1}{|l|}{\multirow{-3}{*}{sf2}} & 3 & 9.271E-03 & 1.039E-02 & \textbf{8.064E-03} \\ \hline
\multicolumn{1}{|l|}{} & 1 & 1.779E-02 & 1.934E-02 & \textbf{1.503E-02}  \\
\multicolumn{1}{|l|}{} & 2 & 1.117E-02 & 1.665E-02 & \textbf{9.304E-03} \\
\multicolumn{1}{|l|}{\multirow{-3}{*}{jura}} & 3 & 1.311E-02 & 1.514E-02 & \textbf{1.262E-02} \\ \hline
\multicolumn{1}{|l|}{} & 1 & \textbf{1.509E-04} & 2.738E-04 & 4.167E-04 \\
\multicolumn{1}{|l|}{\multirow{-2}{*}{enb}} & 2 & \textbf{1.071E-03} & 1.227E-03 & 1.160E-03\\ \hline
\multicolumn{1}{|l|}{} & 1 & 1.622E-01 & \textbf{7.611E-02} & 9.485E-02 \\
\multicolumn{1}{|l|}{} & 2 & 8.823E-02 & 1.050E-01 & \textbf{4.734E-02} \\
\multicolumn{1}{|l|}{\multirow{-3}{*}{slump}} & 3 & 8.423E-03 & \textbf{1.737E-03} & 7.744E-03 \\ \hline
\multicolumn{1}{|l|}{} & 1 & \textbf{1.598E-03} & 1.903E-03 & 2.058E-03 \\
\multicolumn{1}{|l|}{} & 2 & \textbf{1.952E-03} & 2.703E-03 & 2.490E-03 \\
\multicolumn{1}{|l|}{} & 3 & 3.029E-03 & 3.194E-03 & \textbf{2.919E-03} \\
\multicolumn{1}{|l|}{\multirow{-4}{*}{scm1d}} & 4 & \textbf{2.666E-03} & 3.656E-03 & 3.272E-03 \\ \hline
\multicolumn{1}{|l|}{} & 1 & 2.541E-03 & 2.672E-03 & \textbf{2.533E-03} \\
\multicolumn{1}{|l|}{} & 2 & 3.640E-03 & 3.174E-03 & \textbf{3.146E-03} \\
\multicolumn{1}{|l|}{} & 3 & 3.658E-03 & 4.015E-03 & \textbf{3.201E-03} \\
\multicolumn{1}{|l|}{\multirow{-4}{*}{scm20d}} & 4 & 3.756E-03 & 4.115E-03 & \textbf{3.670E-03} \\ \hline
\multicolumn{1}{|l|}{} & 1 & \textbf{2.122E-03} & 2.558E-03 & 2.300E-03 \\
\multicolumn{1}{|l|}{} & 2 & \textbf{3.680E-03} &  4.038E-03  & 3.846E-03 \\
\multicolumn{1}{|l|}{\multirow{-3}{*}{bike}} & 3 & \textbf{3.731E-03} & 4.303E-03  & 3.910E-03 \\ \hline
\multicolumn{1}{|l|}{} & 1 & 2.582E-04 & \textbf{1.317E-04} & 1.518E-04 \\
\multicolumn{1}{|l|}{} & 2 & 1.643E-04 & 8.310E-05 & \textbf{8.307E-05} \\
\multicolumn{1}{|l|}{\multirow{-3}{*}{sarcos}} & 3 & 3.325E-04 & \textbf{1.788E-04} & 1.933E-04 \\ \hline
\multicolumn{1}{|l|}{\multirow{-1}{*}{\# wins}} & - & 14 & 5 & \textbf{23} \\ \hline
\end{tabular}}
\end{table}

\paragraph{Results}
We present results for fully observed response settings in Table \ref{tab:fully-observed-response} and partially observed response settings in Table \ref{tab:partially-observed-response}. In both cases, regularized multi-task soft trees lead the charts over the corresponding baselines in terms of out-of-sample mean squared error performance. For the fully observed response setting, we also show tree ensemble sizes in Table \ref{tab:tree-ensemble-sizes}. We see a large reduction in the number of trees with out proposed multi-task tree ensembles.

\begin{table*}[!htb]
\centering
\caption{Out-of-sample performance of single-task and multi-task tree ensembles with flexible loss functions for zero-inflation/overdispersion. We evaluate performance with weighted Poisson deviance and AUC across tasks.}
\label{tab:out-of-sample-LM}
\small
\resizebox{\textwidth}{!}{\begin{tabular}{lc|c|cccccc|}
\cline{3-9}
\multicolumn{1}{c}{} &  & \textbf{GBDT} & \multicolumn{6}{c|}{\textbf{Soft Trees}} \\ \cline{3-9} 
\multicolumn{1}{c}{} &  & \textbf{Single-task} & \multicolumn{3}{c|}{\textbf{Single-task}} & \multicolumn{3}{c|}{\textbf{Multi-task}} \\ \hline
\multicolumn{1}{|l|}{\textbf{Metric}} & \textbf{Task} & \textbf{Poisson} & \multicolumn{1}{c|}{\textbf{Poisson}} & \multicolumn{1}{c|}{\textbf{ZIP}} & \multicolumn{1}{c|}{\textbf{NB}} & \multicolumn{1}{c|}{\textbf{Poisson}} & \multicolumn{1}{c|}{\textbf{ZIP}} & \textbf{NB} \\ \hline
\multicolumn{1}{|l|}{\multirow{3}{*}{\begin{tabular}[c]{@{}l@{}}Poisson\\ Deviance\end{tabular}}} & 1 & 2.643E-04 & 2.624E-04 & 2.623E-04 & \multicolumn{1}{c|}{2.623E-04} & 2.607E-04 & \textbf{2.605E-04} & 2.608E-04 \\
\multicolumn{1}{|l|}{} & 2 & 8.029E-04 & 8.050E-04 & 8.029E-04 & \multicolumn{1}{c|}{8.044E-04} & 8.022E-04 & \textbf{8.014E-04} & \textbf{8.014E-04} \\
\multicolumn{1}{|l|}{} & 3 & 1.044E-03 & 1.045E-03 & 1.043E-03 &  \multicolumn{1}{c|}{1.042E-03} & 1.041E-03 & \textbf{1.040E-03} & 1.041E-03 \\ \hline
\multicolumn{1}{|l|}{\multirow{3}{*}{AUC}} & 1 &  0.710 & 0.721 & 0.722 & \multicolumn{1}{c|}{0.721} & 0.730 & \textbf{0.734} & 0.727 \\
\multicolumn{1}{|l|}{} & 2 & 0.690  & 0.689 & 0.690 & \multicolumn{1}{c|}{0.688} & 0.691 & 0.691 & \textbf{0.692} \\
\multicolumn{1}{|l|}{} & 3 & 0.684 & 0.683 & 0.686 & \multicolumn{1}{c|}{0.685} & 0.687 & \textbf{0.689} & \textbf{0.689} \\ \hline
\end{tabular}}
\end{table*}

\subsection{Large-scale multi-task data from a multinational financial services company}
\label{sec:expts-lm}
We study the performance of our differentiable tree ensembles in a real-word, large-scale multi-task setting from a multinational financial services company. The system encompasses costs and fees for millions of users for different products and services. The dataset has the following characteristics:
(i) It is a multi-task regression dataset with 3 tasks.
(ii) Each task has high degree of over-dispersion.
(iii) All tasks are not fully observed as each user signs up for a subset of products/services. The degree of missing responses on average across tasks is $\sim 50\%$. 
(iv) Number of features is also large ($\sim600$).

We validate the flexibility of our end-to-end tree-ensemble learning framework with soft trees on a dataset of 1.3 million samples. We study the following flexible aspects of our framework: (i) Flexible loss handling with zero-inflation Poisson regression and negative binomial regression for single-task learning. (ii) Multi-task learning with our proposed regularized multi-task soft tree ensembles in the presence of missing responses across tasks. (iii) Flexible loss handling with zero-inflation Poisson/negative binomial regression in the context of multi-task learning.  

We present our results in Table \ref{tab:out-of-sample-LM}. We can see that we achieve the lowest Poisson deviance and highest AUC with multi-task regression via zero-inflated Poisson/negative binomial regression. 

\section{Conclusion}
We propose a flexible and scalable tree ensemble learning framework with seamless support for new loss functions. Our framework makes it possible to compare different loss functions or incorporate new differentiable loss functions and train them with tree ensembles. Our proposal for tensor-based modeling of tree ensembles allows 10x faster training on CPU than existing toolkits and also allow for training differentiable tree ensembles on GPUs. We also propose a novel extension for multi-task learning with trees that allow soft sharing of information across tasks for more compact and expressive ensembles than those by existing tree ensemble toolkits. 

\subsection*{Acknowledgements}
We thank Denis Sai for his help with the
experiments on large-scale multi-task data. This research was supported in part, by grants from the Office of Naval Research: ONR-N00014-21-1-2841 and award from Liberty Mutual Insurance.

\bibliographystyle{unsrtnat}
\bibliography{references}

\clearpage
\appendix

\section*{Supplementary Material}
\setcounter{table}{0}
\renewcommand{\thetable}{S\arabic{table}}%
\setcounter{figure}{0}
\renewcommand{\thefigure}{S\arabic{figure}}%
\setcounter{equation}{0}
\renewcommand{\theequation}{S\arabic{equation}}
\setcounter{section}{0}
\renewcommand{\thesection}{S\arabic{section}}%
\setcounter{footnote}{0}

\section{Additional details for Experimental Section \ref{sec:experiments}}
\label{supp-sec:experiments}
\paragraph{Datasets} We use a collection of 27 open-source regression datasets from various domains (e.g., social media platforms, human behavior, financial risk data). 9 of these are from Mulan: A Java library for multi-label learning (Mulan) \citep{Xioufis2016}, 2 of them are from University of California Irvine data repository (UCI) \citep{Dua2019}, 12 of them are from Delve database \citep{Akujuobi2017} and the 5 remaining are SARCOS \footnote{The original test set has significant data leakage as noted by \url{https://www.datarobot.com/blog/running-code-and-failing-models/}. Following their guidance we discard the original test set and use the original train set to generate train/validation/test splits.} \citep{Vijayakumar2000}, Youth Risk Behavior Survey\footnote{\url{https://www.cdc.gov/healthyyouth/data/yrbs/data.htm}} \citep{Jacobs2021}, Block-Group level data from US Census Planning Database\footnote{\url{https://www.census.gov/topics/research/guidance/planning-databases.html}}\citep{Bureau2021}, and financial services loss data from Kaggle\footnote{\url{https://bit.ly/3swGnTo}}. For scm1d and scm20d from Mulan\citep{Xioufis2016}, we consider the first 4 tasks (out of the 16 tasks in the original dataset). For SARCOS, we consider 3 torques for prediction (torque-3, torque-4 and torque-7; we ignore the other torques as those seem to have poor correlations with these.)

For all datasets, we split the datasets into $64\%/16\%/20\%$ training/validation/test splits. We train the models on the training set, perform hyperparameter tuning on the validation set and report out-of-sample performance on the test set. 

\subsection{Tuning parameters and optimal depths comparison for a single tree in Section \ref{sec:expts-studying-single-tree}}
\label{supp-sec:studying-single-tree}
\begin{table}[!b]
\small
\centering
\caption{Optimal depth of a \textit{single} axis aligned and oblique decision tree on various regression datasets.}
\label{tab:single-tree-depth}
\setlength{\tabcolsep}{20.0pt}
\begin{tabular}{|l|cc|cc|}
\hline
 & \multicolumn{2}{c|}{\textbf{Axis-Aligned}} & \multicolumn{2}{c|}{\textbf{Oblique}} \\ \hline
\textbf{Data} & \multicolumn{1}{c|}{\textbf{CART}} & \textbf{TAO} & \multicolumn{1}{c|}{\textbf{TAO}} & \textbf{Soft Tree} \\ \hline
abalone & \multicolumn{1}{c|}{5} & 5 & \multicolumn{1}{c|}{4} & \textbf{2} \\ \hline
pumadyn-32nh & \multicolumn{1}{c|}{6} & 6 & \multicolumn{1}{c|}{4} & \textbf{3} \\ \hline
pumadyn-32nm & \multicolumn{1}{c|}{8} & 8 & \multicolumn{1}{c|}{5} & \textbf{4} \\ \hline
pumadyn-32fh & \multicolumn{1}{c|}{\textbf{3}} & \textbf{3} & \multicolumn{1}{c|}{\textbf{3}} & 5 \\ \hline
pumadyn-32fm & \multicolumn{1}{c|}{6} & 6 & \multicolumn{1}{c|}{5} & \textbf{4} \\ \hline
pumadyn-8nh & \multicolumn{1}{c|}{6} & 6 & \multicolumn{1}{c|}{5} & \textbf{3} \\ \hline
pumadyn-8nm & \multicolumn{1}{c|}{9} & 8 & \multicolumn{1}{c|}{7} & \textbf{5} \\ \hline
pumadyn-8fh & \multicolumn{1}{c|}{5} & 5 & \multicolumn{1}{c|}{4} & \textbf{2} \\ \hline
pumadyn-8fm & \multicolumn{1}{c|}{7} & 7 & \multicolumn{1}{c|}{6} & \textbf{3} \\ \hline
cpu & \multicolumn{1}{c|}{10} & 8 & \multicolumn{1}{c|}{\textbf{5}} & \textbf{5} \\ \hline
cpuSmall & \multicolumn{1}{c|}{8} & 8 & \multicolumn{1}{c|}{\textbf{5}} & \textbf{5} \\ \hline
concrete & \multicolumn{1}{c|}{15} & 8 & \multicolumn{1}{c|}{\textbf{5}} & 9 \\ \hline
\end{tabular}
\end{table}

\paragraph{TAO Implementation} We wrote our own implementation of the TAO algorithm proposed in \cite{Carreira-Perpinan2018}. We considered binary trees with TAO for both axis-aligned and oblique trees. In the case of axis-aligned splits, we initialize the tree with CART solution and run TAO iterations until there is no improvement in training objective. In the case of oblique trees, we initialize with a complete binary tree with random parameters for the hyperplanes in the split nodes and use logistic regression to solve the decision node optimization. We run the algorithm until either a maximum number of iterations are reached or the training objective fails to improve. 

\paragraph{Tuning parameters} For CART and axis-aligned TAO, we tune the depth in the range $[2-20]$. We also optimize over the maximum number of iterations in the interval $[20-100]$. For oblique TAO and soft tree, we tune the depth between $2-10$ and the number of iterations between $20-100$. Additionally, for soft tree, we also tune over the learning rates $[1e-5,1e-2]$ with Adam optimizer and batch sizes $\{64,128,256,512\}$. For a fair comparison, we run all $4$ methods for $100$ trials.

\paragraph{Optimal depths}
We make a comparison of optimal depths between CART, TAO (both axis-aligned and oblique) and soft tree. The soft tree finds a smaller optimal depth in majority cases in comparison with its classical counterpart i.e., oblique TAO tree --- See Table \ref{tab:single-tree-depth}. This is hypothesized to be due to the end-to-end optimization done by soft tree as opposed to a local search performed by the TAO algorithm.

\subsection{Tuning parameters for Sections \ref{sec:expts-zero-inflation}}
\label{supp-sec:expts-zero-inflation}
We use HistGBDT from sklearn \cite{sklearn2013} (GBDT in sklearn does not support Poisson regression). We tune over depths in the range $[2-20]$, number of trees between $50-1500$ and learning rates on the log-uniform scale in the interval $[1e-5, 1e-1]$. For differentiable tree ensembles, we tune number of trees in the range $[2,100]$, depths in the set $[2-4]$, batch sizes $\{64,128,256,512\}$, learning rates $[1e-5,1e-1]$ with Adam optimizer and perform early stopping with a patience of $25$ based on the validation set. For all models, we perform a random search with 1000 hyperparameter tuning trials.

\subsection{Tuning parameters for Sections \ref{sec:expts-multi-task}}
\label{supp-sec:expts-multi-task}
We tune number of trees in the interval $[50-1500]$ for RF, GRF and GBDT. For RF and GBDT, we also tune over depths between $2-20$. For GBDT, we tune learning rates between $[1e-5, 1e-1]$. For GRF, we also tune over min\_node\_size in the set $[2-20]$ and $\alpha \in [1e-3, 1e-1]$. For single-task and multi-task trees, we tune over depths $[2-4]$, number of trees $[5-100]$, batch sizes $\{64,128,256,512\}$, epochs $[20-500]$, Adam learning rates $[1e-5, 1e-2]$. We also optimize over the regularization penalty for multi-task soft decision trees $[1e-5, 1e1]$. All single-task models (soft tree ensembles, GBDT) are tuned for 1000 trials per task. All multi-task models (RF, GRF, multi-task soft trees) are tuned for 1000 trials in total.

\subsection{Tuning parameters for Sections \ref{sec:expts-lm}}
\label{supp-sec:expts-lm}
We use GBDT from XGBoost \citep{xgboost2016}, where we tune number of trees in the interval $[50-1500]$, depths between $2-20$ and learning rates between $[1e-4, 1e-0]$. For single-task and multi-task trees, we tune over depths $[2-4]$, number of trees $[5-100]$, batch sizes $\{64,128,256,512\}$, epochs $[20-200]$, Adam learning rates $[1e-5, 1e-2]$. We also optimize over the regularization penalty for multi-task soft decision trees $[1e-5, 1e1]$. All single-task models (soft tree ensembles, GBDT) with Poisson, Zero-Inflated-Poisson, Negative Binomial are tuned for 1000 trials per task. All multi-task soft-trees are tuned for 1000 trials in total.

\end{document}